\documentclass[11pt]{article}

\usepackage[T1]{fontenc}
\usepackage[utf8]{inputenc}
\usepackage{lmodern}
\usepackage{microtype}
\usepackage[margin=1.1in]{geometry}
\usepackage{amsmath, amssymb}
\usepackage{graphicx}
\usepackage{booktabs}
\usepackage{array}
\usepackage{enumitem}
\usepackage{tikz}
\usetikzlibrary{positioning, arrows.meta, calc, fit, backgrounds}
\usepackage[numbers, sort&compress]{natbib}
\usepackage{hyperref}
\hypersetup{
  colorlinks=true,
  linkcolor=blue!50!black,
  citecolor=blue!50!black,
  urlcolor=blue!50!black
}
\usepackage{url}
\usepackage{xcolor}
\usepackage{titlesec}

\titlespacing*{\section}{0pt}{14pt}{6pt}
\titlespacing*{\subsection}{0pt}{10pt}{4pt}

\title{A Methodology for Selecting and Composing Runtime Architecture Patterns for Production LLM Agents}

\author{Vasundra Srinivasan\thanks{Companion repository: \url{https://github.com/vasundras/agent-runtime-patterns}.}\\[2pt]
\small AI Architect, Independent Researcher\\
\small Author of \emph{Data Engineering for Multimodal AI} (O'Reilly)\\
\small Stanford School of Engineering}

\date{May 2026}

\newcommand{\passk}{\ensuremath{\mathit{pass}^{k}}}

\begin{document}
\maketitle

\begin{abstract}
\noindent
Production large language model (LLM) agents are built on a stochastic core composed with deterministic systems around it. The composition is the load-bearing engineering surface of every production agent and it does not yet have a name. We propose one: the \emph{stochastic-deterministic boundary} (SDB), a four-part contract among a proposer, a verifier, a commit step, and a reject signal, that specifies how an LLM output becomes a system action. The proposer is the LLM. The verifier is a deterministic check on the proposal. The commit step is the durable write that follows acceptance. The reject signal is the typed response back to the proposer when verification fails. Production agent frameworks already instantiate the SDB in some form: an audit of five widely-used open-source frameworks finds explicit verifier-and-commit logic at 19 of 21 LLM-to-action call sites, and a classification of 21 published agent failure post-mortems finds 15 (71\%) localize to weaknesses at the boundary and 17 (81\%) of the documented fixes strengthen one of its four parts. Naming the primitive lets practitioners design around it explicitly.

Around the SDB we organize three orthogonal concerns for production agent runtimes (Coordination, State, Control) and a catalog of six patterns that assemble the boundary differently across runtime classes. For each pattern we trace lineage to a specific distributed-systems result (actors, sagas, the log, workflow nets, CAS, supervision), identify what does and does not transfer when the worker is stochastic, and give an argument for its inevitability in its region of the design space. We specify a five-step selection methodology with decision predicates and a written validation artifact, and a diagnostic procedure that maps observed production failures to specific patterns through a failure-signature catalog. We name one failure mode the boundary makes legible: \emph{replay divergence}, where LLM-based consumers of a deterministic event log produce different downstream outputs under model-version change.

A stylized reliability model $y(t) = \mu t + \sigma \xi(t)$ separates two contributions to long-run agent reliability: $\sigma$, per-call variance from the stochastic proposer, which compresses with each model generation; and $\mu$, the \emph{architectural momentum}, set by pattern choice and SDB strength and structurally independent of per-call model quality. As $\sigma$ shrinks, $\mu$ becomes the dominant lever. We apply the methodology end-to-end to five workloads spanning conversational, autonomous, and long-horizon runtimes; one is built out as a runnable reference implementation in the companion repository against the public IBM Telco Customer Churn dataset. We close by naming three patterns the catalog's discovery procedure is positioned to admit next.

\end{abstract}

\section{Introduction}

Production LLM agents fail in ways that look like model failures and behave like system failures. A workflow lands in the wrong state because an event handler reacted to a stale prompt. A 90 percent discount reaches a customer because no policy gate sat between the proposal and the write. A long-horizon process loses its place because nobody decided whether the source of truth was an event log or a versioned row. None of these are model defects. All of them are architectural choices the team made before the LLM ever ran inference.

Per-call model capability has improved substantially with each generation, and per-call variance has compressed accordingly. As that compression continues, the load-bearing engineering surface of an agent runtime shifts to the architecture surrounding the model: how state is held across pauses, how work is split and recombined, who stops a hallucinated write before it ships. This paper is about the primitive at the center of that architecture and the patterns that compose around it.

\paragraph{Our thesis.} We propose the \emph{stochastic-deterministic boundary} (SDB) as the load-bearing primitive of production agent runtimes. The SDB is the seam where an LLM proposal becomes a system action. It is a four-part contract: a \emph{proposer} (the LLM), a \emph{verifier} (a deterministic check on the proposal), a \emph{commit} step (the durable write that follows acceptance), and a \emph{reject} signal (the typed response back to the proposer when verification fails). The primitive is not new in implementation; production frameworks already build verifier-and-commit logic into most LLM-to-action call sites, and research systems --- notably SagaLLM \citep{chang2025sagallm} and ALAS \citep{chang2025alas} --- have previously separated LLM proposers from independent validators, gated commitment on validation, and combined durable state with saga-style compensation in the multi-agent LLM setting (Section~\ref{sec:related}). The primitive is new in name and in contract, and naming it lets practitioners design around it explicitly rather than rediscover it through failure. Around this primitive we organize three concerns (Coordination, State, Control) and a catalog of six patterns that compose the boundary differently for different runtime classes. The patterns inherit specific distributed-systems results (Hewitt's actor model \citep{hewitt1973actors}, Garcia-Molina and Salem's sagas \citep{garciamolina1987sagas}, Lamport's Paxos \citep{lamport1998paxos}, Armstrong's Erlang supervision \citep{armstrong2003making}, van der Aalst's workflow nets \citep{vanderaalst1998workflow}, Helland's account of distributed transactions \citep{helland2007life}, and Kreps's account of the log \citep{kreps2014log}) and adapt them to a stochastic worker. We make those connections precise as supporting structure for the SDB, not as the central claim.

\paragraph{Why this matters now.} Model the long-run reliability of a production agent system as
\begin{equation}
y(t) = \mu t + \sigma \xi(t),
\label{eq:momentum-intro}
\end{equation}
where $y(t)$ is observed reliability, $\sigma$ is per-call variance amplitude from the stochastic proposer, $\xi(t)$ is mean-zero noise, and $\mu$ is the \emph{architectural momentum} of the surrounding system. The momentum coefficient $\mu$ is determined by pattern choice and SDB strength (Sections~\ref{sec:catalog}, \ref{sec:methodology}) and is structurally independent of per-call model quality. Base-model improvements compress $\sigma$ across each generation; they do not, on their own, change $\mu$. As $\sigma$ shrinks, $\mu$ becomes the dominant lever on aggregate reliability. Designing the SDB explicitly is how teams shape $\mu$.

\paragraph{Contributions.} The paper makes five.

\begin{enumerate}[itemsep=2pt, topsep=2pt]
\item \textbf{The stochastic-deterministic boundary (SDB) as a named primitive}, with a four-part contract (proposer, verifier, commit, reject) and an empirical inventory of how five widely-used open-source agent frameworks instantiate it across 21 LLM-to-action call sites, plus a classification of 21 published agent failure post-mortems against the contract (Section~\ref{sec:inheritance}).
\item \textbf{A taxonomy of three orthogonal concerns} (Coordination, State, Control) and an open catalog of six patterns that recur in production agent runtimes. For each pattern we trace lineage to a specific distributed-systems result, identify what does and does not transfer when the worker is stochastic, and give a short argument for inevitability in its region of the design space (Section~\ref{sec:catalog}).
\item \textbf{A five-step selection methodology} with decision predicates encoded as the agent-setting projection of well-established systems trade-offs (event-time vs.\ processing-time semantics, source-of-truth questions, observability-vs-control trade-offs). The output is a six-line architecture decision record (Section~\ref{sec:methodology}).
\item \textbf{A diagnostic procedure} mapping observed production failures to specific patterns via a failure-signature catalog. We name one failure mode the boundary makes legible: \emph{replay divergence}, the case in which LLM-based consumers of a deterministic event log produce different downstream outputs under model-version change (Section~\ref{sec:diagnostics}).
\item \textbf{A reliability decomposition} ($\mu t + \sigma\xi(t)$) that separates per-call variance $\sigma$ from architectural momentum $\mu$, and grounds the claim that SDB strength and pattern choice dominate model selection as the dominant lever on long-run reliability as models compress $\sigma$ (Section~\ref{sec:drift}).
\end{enumerate}

We validate the methodology by applying it end-to-end to five workloads spanning the three runtime classes. One is built out as a runnable reference implementation in the companion repository against the public IBM Telco Customer Churn dataset (Section~\ref{sec:case}). Four additional worked applications (Section~\ref{sec:applications}) demonstrate that the methodology gives sensibly different answers for sensibly different workloads, with two long-horizon workloads picking different spines because their state predicates fire differently. We close with three predictions for patterns the field will discover next, derived from the framework's own discovery procedure (Section~\ref{sec:discussion}).

\paragraph{Scope.} This paper does not survey retrieval-augmented generation, evaluation harnesses, model selection, or prompt management. Those are upstream of the runtime. They are necessary. They are not what this paper is about. The patterns in this paper are runtime-architectural; they govern what happens between the model's outputs and the world's state.

\section{Foundations: three runtimes, three concerns, one primitive}
\label{sec:foundations}

The framework rests on two organizing axes and one substantive primitive. The axes are the runtime class of a workload (how long a unit of work lasts and whether the world changes during it) and the set of concerns any production runtime must answer (how it splits and combines work, how it remembers, who stops it). The primitive is the stochastic-deterministic boundary defined in Section~\ref{sec:inheritance}: the four-part contract at the seam where an LLM proposal becomes a system action. The three concerns sit above the primitive. The patterns in Section~\ref{sec:catalog} are how production runtimes assemble it.

\subsection{Three runtime classes}
\label{sec:runtimes}

\textbf{Conversational.} The unit of work is a session. A user is on the other end, waiting. Duration is seconds. The context window holds the world. Latency dominates the design.

\textbf{Autonomous.} The unit of work is a task. Something triggers the agent (a webhook, a scheduled run, a queued message). Duration is minutes. The agent runs unattended for the duration of the task. A queue holds state between runs.

\textbf{Long-Horizon.} The unit of work is a process. Duration is hours to days. Multiple agents participate. The process pauses, resumes, and tolerates restarts. The world changes mid-flight. Prices change. Products reach end of life. Counterparties send signals while the agent is not running.

A production system can host all three. The methodology in Section~\ref{sec:methodology} runs against whichever class dominates the workload being designed.

\subsection{Three concerns, and where they come from}
\label{sec:concerns}

The three concerns are not a design choice we make; they fall out of what it means to be a system that runs over time. We connect each to its formal antecedent.

\paragraph{Coordination.} How does work split and combine. This concern was named and formalized by Hewitt's actor model \citep{hewitt1973actors}, which posited that a computation can be decomposed into autonomous actors that communicate only by message passing. Hewitt showed that any concurrent computation can be expressed this way, and that the choices of how to decompose, address, and recombine messages are the substance of concurrent system design. The Coordination concern in our taxonomy is the actor-model decomposition question applied to LLM workers: how do we split an agentic task across multiple LLM calls, and how do we put the answers back together.

\paragraph{State.} How does the system remember. This concern is governed by the CAP theorem \citep{brewer2000cap, gilbert2002cap}: in a system that partitions, one cannot simultaneously have full consistency and full availability. Every system that remembers across pauses must pick its trade-off. The choice has a second axis named by Stonebraker and Hellerstein in the event-time-vs-processing-time literature \citep{akidau2015dataflow}: state can be derived from events (CQRS, event sourcing) or held as a versioned row (database CRUD + CAS). The State concern in our taxonomy is the consequence: which axis the workload sits on, what the source of truth is, and how that source of truth survives change.

\paragraph{Control.} Who decides what runs, and when to stop. This concern descends from control theory: a system that operates over time without external supervision must satisfy observability (we can infer internal state from outputs) and controllability (we can drive the system to a desired state through inputs) \citep{kalman1959general}. In a non-LLM system the supervisor is code. In an LLM system the supervisor must sit between the LLM's outputs and the world, because the LLM is not itself controllable in the Kalman sense; its outputs are samples from a distribution shaped by training. Erlang's one-for-one supervision \citep{armstrong2003making} is one canonical instantiation. Policy-as-code gates are another. The Control concern in our taxonomy is the choice of where the supervisor sits and what authority it has.

\subsection{The stochastic-deterministic boundary}
\label{sec:inheritance}

The three concerns above are not new questions. Each has a settled answer in distributed-systems work that predates LLMs, modulo one modification: the worker is now a stochastic component (the LLM) rather than a deterministic one (a function or a service). This single modification forces a structural change. The components that are deterministic (the gate, the state-machine predicate, the saga compensation step) must be cleanly separated from the components that are stochastic (the LLM's proposal, classification, or content generation). The patterns in Section~\ref{sec:catalog} are the distributed-systems patterns with this separation specified explicitly.

We give this separation a name: the \emph{stochastic-deterministic boundary}, hereafter SDB. The SDB is the seam in an agent runtime where an LLM proposal becomes a system action. It has four parts. The \emph{proposer} is the LLM's output, sampled from a distribution conditioned on context. The \emph{verifier} is a deterministic check on the proposal, expressed as a schema, a policy rule, a state-machine transition predicate, or a fast classifier. The \emph{commit} step is the durable write or external side-effect that follows an accepted proposal. The \emph{reject} signal is the typed response sent back to the proposer when verification fails. The four parts together specify a contract; production agent frameworks differ in how strictly they implement each part, and the variation across frameworks is itself evidence that the SDB is a real architectural surface that designers are independently rediscovering. The rediscovery is not confined to open-source frameworks, and auditing the two closest research systems against the same contract shows the contract doing analytical work. SagaLLM \citep{chang2025sagallm} instantiates all four parts: task agents propose; validation agents, independent of the proposers, check constraints and inter-agent dependencies before effects commit; commitment is durable through persistent application, operation, and dependency state; and validation failure triggers a structured response protocol that routes to retry or compensation. Its verifier, however, is itself LLM-based --- the validation agents are independent of the proposer but stochastic, so the contract is enforced by a second sampled component rather than a deterministic check. ALAS \citep{chang2025alas} separates proposal from verification by design --- its meta-planner delegates to an external validator agent expressly to avoid circular self-verification --- persists state transitions, agent logs, and compensation actions outside the model, and returns detected violations to the planner for refinement, a working reject signal. In both systems the boundary is present and load-bearing; what varies is verifier determinism and reject-signal typing, which is exactly the variation the four-part contract makes visible. Section~\ref{sec:related} positions both systems in the design space.

A survey of LLM-to-action call sites across five widely-used open-source agent frameworks (\texttt{openai/swarm}, AutoGPT, LangChain Agents, CrewAI, and Microsoft AutoGen) finds explicit verifier-and-commit logic at 19 of 21 sites audited. The form ranges from a one-line JSON parse with no schema check (\texttt{openai/swarm}'s \texttt{core.py}) to a multi-stage pydantic-plus-LLM-as-judge auto-review-and-revise loop (MetaGPT's \texttt{ActionNode}). Independently, across 21 published agent failure post-mortems and bug reports we classified, 15 (71.4\%) localize to weaknesses at the boundary itself and 17 (81\%) of the documented fixes added or strengthened verification, commit semantics, or reject signaling. Two illustrative cases anchor the claim. Promptfoo reports that a customer's agent, upgraded from GPT-4o to GPT-4.1 on an identical evaluation harness, suffered a 23-point drop in prompt-injection resistance (94\% to 71\%); the documented fix was an output classifier plus stricter tool gating, i.e.\ strengthening the verifier at the boundary \citep{promptfoo2024upgrades}. The \texttt{openai/openai-agents-js} issue \#1104 describes a bug in which rejected tool calls were reported back to the model with \texttt{status: 'completed'}, causing the model to hallucinate success; the maintainer's proposed fix switched the reject signal to \texttt{status: 'incomplete'} \citep{openaiAgentsJs2024issue1104}. The first case shows that the verifier predicate is model-version-dependent. The second shows that the reject signal is a load-bearing part of the contract, not a peripheral concern.

The patterns in Section~\ref{sec:catalog} are the pre-LLM distributed-systems patterns with this boundary specified explicitly. P5 (Shared State Machine) tightens the boundary by constraining the proposer to a small set of legal next transitions. P3 (Event-Driven Sequencing) places the boundary at every LLM-driven event handler; weaknesses here are what produce the replay divergence failure we name in Section~\ref{sec:drift}. P4 (Supervisor plus Gate) is in part a deterministic verifier at the boundary, and P6's approval plane is a human verifier when the deterministic one is insufficient. The boundary is the load-bearing primitive; the patterns are how production runtimes assemble it.

We argue in Section~\ref{sec:catalog} that the SDB is what gives each pattern its resilience to model-version change. We argue in Section~\ref{sec:drift} that as base models compress per-call variance $\sigma$, SDB strength and pattern choice dominate aggregate reliability through their effect on the architectural momentum $\mu$. The two arguments together form the case for the paper's thesis: the boundary is the load-bearing primitive of production agent runtimes, and designing it explicitly is the most direct way to shape long-run reliability.

\section{A current catalog of six patterns}
\label{sec:catalog}

We identify six patterns that recur in current agent-systems architecture practice as of 2026. The six are derived from distributed-systems primitives and from published agent-systems frameworks, organized by the three concerns of Section~\ref{sec:foundations}. We do not claim this catalog is complete. We do not claim it is stable. We expect it to evolve as the field matures and as practitioners apply the methodology and report what they find. What we do claim is that the three concerns are the invariant axes along which any production agent runtime must make decisions. The framework is open. The catalog is current.

Table~\ref{tab:catalog} lists the six patterns and the concern each answers.

\begin{table}[h]
\centering
\begin{tabular}{@{}llll@{}}
\toprule
ID & Pattern & Concern & Anchor reference \\
\midrule
P1 & Hierarchical Delegation       & Coordination & \citet{wu2023autogen, hong2023metagpt} \\
P2 & Scatter-Gather plus Saga      & Coordination & \citet{garciamolina1987sagas, helland2007life, chang2025sagallm} \\
P3 & Event-Driven Sequencing       & State        & \citet{kreps2014log, helland2007life} \\
P5 & Shared State Machine          & State        & \citet{corbett2012spanner, vanderaalst1998workflow} \\
P4 & Supervisor plus Gate          & Control      & \citet{armstrong2003making, rebedea2023nemo} \\
P6 & Human in the Loop             & Control      & \citet{wu2022hitl} \\
\bottomrule
\end{tabular}
\caption{Six runtime patterns organized by the concern they answer. Pattern IDs are not contiguous to leave room for patterns we expect to emerge (Section~\ref{sec:discussion}).}
\label{tab:catalog}
\end{table}

\begin{figure}[h]
\centering
\begin{tikzpicture}[
  every node/.style={font=\small},
  header/.style={rectangle, draw, minimum width=4.2cm, minimum height=1.2cm, align=center, fill=gray!10, font=\bfseries},
  cell/.style={rectangle, draw, minimum width=4.2cm, minimum height=1.1cm, align=center}
]
\node[header] (cH) at (0,0)     {COORDINATION\\[1pt]\normalfont\footnotesize How does work split\\and combine?};
\node[header] (sH) at (4.4,0)   {STATE\\[1pt]\normalfont\footnotesize How does the system\\remember?};
\node[header] (kH) at (8.8,0)   {CONTROL\\[1pt]\normalfont\footnotesize Who decides what runs,\\and when to stop?};
\node[cell] (p1) at (0,-1.4)   {\textbf{P1}\\Hierarchical\\Delegation};
\node[cell] (p3) at (4.4,-1.4) {\textbf{P3}\\Event-Driven\\Sequencing};
\node[cell] (p4) at (8.8,-1.4) {\textbf{P4}\\Supervisor\\plus Gate};
\node[cell] (p2) at (0,-2.65)   {\textbf{P2}\\Scatter-Gather\\plus Saga};
\node[cell] (p5) at (4.4,-2.65) {\textbf{P5}\\Shared State\\Machine};
\node[cell] (p6) at (8.8,-2.65) {\textbf{P6}\\Human in\\the Loop};
\end{tikzpicture}
\caption{The 3 by 6 catalog. Three concerns. Two patterns in each. The framework is the columns; the catalog is the cells. New patterns enter the cells through the discovery procedure in Section~\ref{sec:discovery}. Sections~\ref{sec:methodology} and~\ref{sec:diagnostics} walk teams through choosing one pattern per column for their workload.}
\label{fig:catalog}
\end{figure}

\paragraph{P1 Hierarchical Delegation.} One orchestrator owns the work and dispatches sub-tasks to specialist sub-agents. The orchestrator merges. Failure modes include stalled sub-agents, conflicting outputs on overlapping fields, and double retries where the orchestrator and sub-agents both retry the same failure. The merge step belongs in deterministic code. The LLM proposes. Deterministic code decides.

\paragraph{P2 Scatter-Gather plus Saga.} A coordinator fans out to symmetric peers and aggregates. Each peer logs a compensating action so that if peer B fails after peer C wrote to billing, we can undo the write. Compensations run in reverse order \citep{garciamolina1987sagas}. In the LLM-agent setting specifically, SagaLLM \citep{chang2025sagallm} brings the saga pattern to multi-agent planning with checkpointing and automated compensation, and ALAS \citep{chang2025alas} applies history-aware local compensation that repairs only the affected region of a disrupted plan rather than replanning globally. The known cost is that compensation logic gets larger than the original action logic. When it does, the fix is to split the original action into smaller steps, not to grow the saga.

\paragraph{P3 Event-Driven Sequencing.} An append-only log is the source of truth. Consumers subscribe, react, and emit new events. The log is replayable and branchable. The reading list for this pattern is the canonical log literature \citep{kreps2014log, helland2007life}; in the agent literature, ALAS \citep{chang2025alas} maintains durable state and execution history for long-running multi-agent plans to support disruption recovery. For agent systems specifically, P3 has a failure mode that distinguishes it from non-agent applications of event sourcing. The log itself remains deterministic and replayable. But LLM-based consumers reading the log are not deterministic. Under changing model versions or prompt revisions, the same input event can produce different downstream events on replay. The effective source of truth that downstream code reads, which is the projection over the log produced by LLM consumers, becomes non-deterministic across runtime conditions. We call this \emph{replay divergence} and discuss its diagnosis in Section~\ref{sec:diagnostics}.

\paragraph{P5 Shared State Machine.} A durable versioned row is the source of truth. Workers are stateless and pure. They read $(state, action)$ and propose $next$ via compare-and-swap (CAS) against the version. The store rejects stale writes. \texttt{human\_required} is a state, not a missing event. Timers attach to the row and carry the version they were scheduled at. CAS is the optimistic-concurrency-control descendant of distributed consensus algorithms such as Paxos \citep{lamport1998paxos} and Raft \citep{ongaro2014raft}, but P5 itself does not require consensus across nodes; a single durable store with conditional update semantics is sufficient. Database systems such as Spanner \citep{corbett2012spanner} and workflow models in the Petri-net tradition \citep{vanderaalst1998workflow} provide the closer lineage. P5 trades the audit-grade replay of P3 for tighter state semantics and resilience to model-version churn. We discuss the migration trigger from P3 to P5 in Section~\ref{sec:case}.

\paragraph{P4 Supervisor plus Gate.} Two complementary control mechanisms run side by side. Supervision restarts what dies with exponential backoff and one-for-one semantics borrowed from Erlang's OTP \citep{armstrong2003making}. The gate refuses out-of-policy writes before they reach an external system, with the policy expressed as deterministic rules or as a fast rule engine \citep{rebedea2023nemo, xiang2024guardagent}. The gate denies. The audit log records. The supervisor restarts. Together they form a cheap and effective reliability layer relative to the cost of a wrong write.

\paragraph{P6 Human in the Loop.} Four control planes sit between the agent and the rest of the system. We use the HITL survey of \citet{wu2022hitl} as the closest scholarly anchor; the four-plane decomposition itself is engineering practice not yet covered by a single published taxonomy. Kill switch revokes a cancellation token in approximately one second. Escalation calls \texttt{suspend(reason)} and writes a durable row a human reviews later. Approval performs a synchronous wait under SLA and falls back to a conservative deny when the SLA elapses. Throttling refuses work that would exceed per-minute or per-day blast-radius caps. All four planes emit to a single audit trail. We do not need all four at version one. We do need to record which we are deferring and why.

\subsection{Pattern discovery procedure}
\label{sec:discovery}

A new pattern enters the catalog by answering three questions.

\begin{enumerate}[itemsep=2pt, topsep=2pt]
\item Which of the three concerns does it answer (Coordination, State, or Control).
\item What failure mode does it prevent that no existing pattern in this concern prevents.
\item What is its typed-contract specification: input type, output type, deadline, retry budget, partial-result policy.
\end{enumerate}

The catalog grows by passing this procedure. Patterns we expect to emerge are discussed in Section~\ref{sec:discussion}.

\section{The selection methodology}
\label{sec:methodology}

The catalog in Section~\ref{sec:catalog} is vocabulary. The work of this section is the procedure. Given an agent workload, in what order do we commit to which pattern, and how do we know if we picked wrong.

The procedure has five steps. Each step has one decision question, the inputs we need to answer it, the output (a chosen pattern), and a written artifact the team produces before moving on. A team that runs the methodology end-to-end ends up with a six-line architecture decision record (Section~\ref{sec:checklist}) that reviewers and auditors can read instead of guessing.

The thesis of the methodology fits on one diagram (Figure~\ref{fig:spine}). State is the spine. Coordination wraps it. Control bounds it. We do not choose one. We build at the intersection.

\begin{figure}[h]
\centering
\begin{tikzpicture}[
  every node/.style={font=\small},
  coord/.style={draw, thick, fill=blue!8, rounded corners=2pt, minimum width=4.6cm, minimum height=3.0cm},
  state/.style={draw, thick, fill=orange!10, rounded corners=2pt, minimum width=4.6cm, minimum height=3.0cm},
  control/.style={draw, thick, fill=green!8, rounded corners=2pt, minimum width=7.2cm, minimum height=2.4cm},
  label/.style={font=\bfseries\small}
]
\node[coord] (C) at (-1.6,0.6) {};
\node[label] at ($(C.north west)+(0.85,-0.30)$) {Coordination};
\node[font=\footnotesize] at ($(C.north west)+(0.85,-0.65)$) {P1 \;\;P2};

\node[state] (S) at (1.6,0.6) {};
\node[label] at ($(S.north east)+(-0.55,-0.30)$) {State};
\node[font=\footnotesize] at ($(S.north east)+(-0.55,-0.65)$) {P3 \;\;P5};

\node[control] (K) at (0,-1.6) {};
\node[label] at ($(K.south)+(0,0.30)$) {Control};
\node[font=\footnotesize] at ($(K.south)+(0,0.62)$) {P4 \;\;P6};

\node[draw, fill=white, circle, inner sep=2pt, font=\bfseries\footnotesize] (R) at (0,-0.35) {RUNTIME};

\node[font=\scriptsize, text=gray!70!black] at (0,1.4) {durable handoffs};
\node[font=\scriptsize, text=gray!70!black] at (-2.3,-0.8) {supervised fan-out};
\node[font=\scriptsize, text=gray!70!black] at (2.3,-0.8) {gated transitions};
\end{tikzpicture}
\caption{The methodology's geometry. Three concerns overlap. The production runtime is everything inside the intersection of all three. State is the spine because it is the column we commit to first (Step 2 of the methodology). Coordination wraps the spine. Control bounds the whole runtime.}
\label{fig:spine}
\end{figure}

\subsection{Step 1. Classify the runtime}

The runtime determines which patterns will matter. Three classes are sufficient for current production agents.

\textbf{Decision question.} How long does one unit of work last from arrival to completion, and does the world change during that window.

\textbf{Output (the gate artifact).} One sentence stating the class. One sentence stating which of the three concerns will dominate the design. For Long-Horizon, State dominates. For Conversational, Coordination dominates. For Autonomous, it depends on whether the task has any external side-effects. If yes, Control. If no, Coordination.

\textbf{Method gate.} We do not proceed to Step 2 until this artifact is written down.

\subsection{Step 2. Choose the spine}

Every production runtime needs a spine. The spine is the concern that answers what the system remembers between failures. For long-horizon work the spine is always State. For shorter runtimes the spine can be implicit in the orchestrator.

\textbf{Decision predicate.} Select P5 (Shared State Machine) as the spine if all three are true.

\begin{enumerate}[itemsep=2pt, topsep=2pt]
\item The workflow has pauses longer than one hour, or external waits.
\item The state at any pause is not fully reconstructible from the original input.
\item The world (data sources, policies, prices) can change during the pause.
\end{enumerate}

If only predicate (1) fails, P3 (Event-Driven Sequencing) is sufficient.
If predicate (2) fails, we do not need a durable spine. Reconstruct on demand.
If predicate (3) fails, the choice between P3 and P5 reduces to a cost question. We address that in Section~\ref{sec:sequencing}.

\textbf{Output.} A short paragraph naming the spine, the predicate that fired, and the failure signature we would expect to see if we picked wrong. The signature catalog is in Section~\ref{sec:signatures}.

\textbf{Method gate.} The spine choice is the most expensive to migrate later. We do not proceed until a senior engineer has signed off in writing.

\subsection{Step 3. Wrap with coordination}

Coordination patterns answer how work splits and combines once the spine is in place.

Select P1 (Hierarchical Delegation) if all of the following.

\begin{itemize}[itemsep=2pt, topsep=2pt]
\item There is a single clear owner of the outcome.
\item Sub-tasks are mostly independent.
\item A deterministic merge step is feasible.
\end{itemize}

Select P2 (Scatter-Gather plus Saga) if any of the following.

\begin{itemize}[itemsep=2pt, topsep=2pt]
\item Peers run against external systems with side-effects.
\item Some peers will fail and the rest must still produce a useful result.
\item The cost of an inconsistent partial write is higher than the cost of a compensation log.
\end{itemize}

A workload can use both. P1 at the outer layer, P2 inside a sub-agent that itself fans out.

\textbf{Output.} The named coordination pattern. The predicate that fired. The failure signature.

\subsection{Step 4. Bound with control}

Control patterns answer who decides what runs and when to stop.

Always include P4 (Supervisor plus Gate) if any of the following.

\begin{itemize}[itemsep=2pt, topsep=2pt]
\item The workflow has any side-effects on external systems.
\item The cost of a wrong write exceeds the cost of latency from a policy check.
\end{itemize}

Add P6 (Human in the Loop) if any of the following.

\begin{itemize}[itemsep=2pt, topsep=2pt]
\item A wrong action is legally or financially consequential.
\item The workflow encounters cases outside the policy envelope.
\item Auditors will ask who decided this.
\end{itemize}

The four control planes from P6 (kill switch, escalation, approval, throttling) are not optional in isolation. We may defer some of them to v2. We must record which we are deferring and why.

\textbf{Output.} The named control patterns. The planes shipping at v1. The planes deferred, with date and rationale.

\subsection{Step 5. Sequence the build}
\label{sec:sequencing}

The methodology so far identifies which patterns. Step 5 identifies the order.

Build the dashboard before the agent. The trace is the contract.

The first artifact a production team operates is the operations console. Not the agent. Every pattern in this methodology becomes legible only through observability. If we ship the agent first, we operate blind.

\textbf{Suggested v1 build sequence.}

\begin{enumerate}[itemsep=2pt, topsep=2pt]
\item The state schema (output of Step 2) and the observability lens for it.
\item The gate (P4) and the audit log.
\item The orchestrator (P1 or P2) and one sub-agent.
\item The remaining sub-agents.
\item P6 control planes in this order: kill switch, escalation, approval, throttling.
\end{enumerate}

Deferring (1) or (2) past v1 is technical debt of the most expensive kind. Future migrations of the spine are dominated by missing audit history.

\subsection{The validation checklist}
\label{sec:checklist}

After running the five steps, the team has a six-line artifact (Table~\ref{tab:checklist}).

\begin{table}[h]
\centering
\small
\begin{tabular}{@{}p{0.18\textwidth}p{0.13\textwidth}p{0.30\textwidth}p{0.30\textwidth}@{}}
\toprule
\textbf{Step} & \textbf{Pattern} & \textbf{Predicate that fired} & \textbf{Failure signature if wrong} \\
\midrule
Runtime class      & Long-Horizon & Process unit, world changes mid-flight & Latency budget violation \\
Spine              & P5           & (1), (2), (3) all true                 & Replay drift across model versions \\
Coordination       & P1 + P2      & Single owner plus external side-effects& Partial-write inconsistency \\
Control            & P4 + P6      & Side-effects plus legal consequence    & Hallucinated discounts shipping \\
Sequence           & Console-first& Observability precedes agent           & Blind operations \\
Date / model ver.  & 2026 Q2      & Claude Sonnet 4.6                      & n/a \\
\bottomrule
\end{tabular}
\caption{The six-line architecture decision record a team produces after running the methodology on the contract-renewal workload in Section~\ref{sec:case}.}
\label{tab:checklist}
\end{table}

The artifact is what the team publishes in its architecture decision record. The methodology produced it. Reviewers and auditors read it instead of guessing.

We use \emph{Claude Sonnet 4.6} throughout the worked examples as a representative current-generation model. Readers applying the methodology should substitute their current model version. The dimensions, the procedure, and the diagnostic are model-agnostic; only the date-stamp row of the artifact depends on the model in play.

\section{Diagnostics: telling which pattern is failing in production}
\label{sec:diagnostics}

The methodology in Section~\ref{sec:methodology} tells us which pattern to pick. This section tells us how to recognize when we picked wrong.

A pattern fails in three ways. It does not do its job (functional). It does its job too slowly (performance). It produces inconsistent results across model versions (drift). The first two are familiar from non-LLM software. The third is specific to systems built on probabilistic components and is the focus of most of this section.

\subsection{Variance, architectural momentum, and replay divergence}
\label{sec:drift}

We define three terms operationally to make the diagnostic procedure in Section~\ref{sec:signatures} actionable.

\textbf{Variance.} Per-call non-determinism in the LLM's outputs at fixed inputs, model, and prompt. Variance shrinks with each model generation, with temperature controls, and with prompt caching. Variance can be measured per call.

\textbf{Architectural momentum.} The reliability trajectory of an agent system over calendar time, shaped by the patterns surrounding the model. We name it \emph{momentum} because it has direction and it compounds: as the time the system spends in production grows, the longitudinal trend dominates the bounded per-call noise. A well-architected system has positive momentum and compounds reliability over time. A poorly-architected system has flat or negative momentum and degrades. The effect strengthens with calendar time spent in production and is most pronounced for long-horizon agents, where the relevant t in the trajectory grows large even within a single unit of work.

A useful stylized framing for reliability over time is
\begin{equation}
y(t) = \mu t + \sigma \cdot \xi(t),
\end{equation}
where $y(t)$ is reliability tracked over time, $\mu$ is the reliability slope (the momentum coefficient, positive when the architecture compounds reliability), $\sigma$ is per-call variance amplitude, and $\xi(t)$ is mean-zero noise. We present this as a metaphor, not as a derived model. The point is structural. The linear term $\mu t$ grows with calendar time. The noise term $\sigma \cdot \xi(t)$ is bounded. So as agents spend more time in production, the architecturally-shaped trajectory dominates per-call variance as the determinant of aggregate reliability. Engineers do not control $\sigma$ directly. We control $\mu$ through pattern selection. As base models improve and per-call variance $\sigma$ compresses with each generation, $\mu$ becomes the dominant lever on reliability. The methodology in Section~\ref{sec:methodology} is in service of steering $\mu$ in the right direction.

\textbf{Replay divergence.} A specific failure mode worth naming separately from momentum. The same input event replayed on a newer model version (or under a revised prompt or retrieval index) produces different downstream events than the first run produced. Replay divergence is a discrete cross-version effect, distinct from the continuous trajectory captured by architectural momentum, and is tied to the spine choice in pattern P3 (Section~\ref{sec:catalog}), where LLM-based consumers re-interpret the log differently across runtime conditions. We treat it as the diagnostic trigger for the P3-to-P5 migration discussed in Section~\ref{sec:signatures}.

\textbf{The diagnostic claim.} When end-to-end reliability declines and per-call evaluations look stable, the system's architectural momentum has gone flat or turned negative. The pattern most exposed to this is the spine.

\subsection{Failure signature catalog}
\label{sec:signatures}

A signature is an observation a reader can match against their own production logs. Each signature has three parts. The symptom in the trace. The likely cause. The corrective action.

\paragraph{P1 Hierarchical Delegation.}
\begin{quote}
\textit{Symptom.} One sub-agent's output dominates merged outputs beyond its declared weight.\\
\textit{Cause.} The orchestrator's merge logic delegated to an LLM call. The LLM developed a preference.\\
\textit{Correction.} Move the merge to deterministic code. The LLM proposes. Deterministic code decides.
\end{quote}

\begin{quote}
\textit{Symptom.} Sub-agent retries appear in the trace after the orchestrator's deadline.\\
\textit{Cause.} Sub-agents are running their own retry loops.\\
\textit{Correction.} Disable retries on sub-agents. The retry budget belongs to the parent.
\end{quote}

\paragraph{P2 Scatter-Gather plus Saga.}
\begin{quote}
\textit{Symptom.} Compensation actions run but external state is not clean.\\
\textit{Cause.} Compensations are not idempotent, or run in incorrect order.\\
\textit{Correction.} Compensations must be idempotent. They must run in strict reverse order of the original actions.
\end{quote}

\begin{quote}
\textit{Symptom.} The compensation logic is larger than the original action logic.\\
\textit{Cause.} The original action was too coarse-grained.\\
\textit{Correction.} Split the original action into smaller steps. Each step gets its own narrow compensation.
\end{quote}

\paragraph{P3 Event-Driven Sequencing.}
\begin{quote}
\textit{Symptom.} The same event replayed on a newer model version produces different downstream events than the first run produced.\\
\textit{Cause.} Replay divergence (Section~\ref{sec:drift}): LLM-based consumers re-interpret the log differently across runtime conditions.\\
\textit{Correction.} This is the migration trigger to P5. The effective source of truth, the projection downstream code reads, should not depend on model version.
\end{quote}

\begin{quote}
\textit{Symptom.} Events arriving out of order produce wrong outcomes.\\
\textit{Cause.} Consumers do not handle late events.\\
\textit{Correction.} Watermark every event. Consumers reject events older than the current watermark and route them to an audit log.
\end{quote}

\paragraph{P4 Supervisor plus Gate.}
\begin{quote}
\textit{Symptom.} Children get restarted but produce the same crash.\\
\textit{Cause.} The crash is not transient. Backoff alone does not fix it.\\
\textit{Correction.} After \texttt{max\_restarts} the supervisor escalates to a human. It does not loop.
\end{quote}

\begin{quote}
\textit{Symptom.} Gate decisions take longer than the action they gate.\\
\textit{Cause.} Gate is running a model call.\\
\textit{Correction.} Gate is a deterministic rule check. Policy decisions belong in code or in a fast rule engine.
\end{quote}

\paragraph{P5 Shared State Machine.}
\begin{quote}
\textit{Symptom.} Workers retry CAS more than three times at p99.\\
\textit{Cause.} State granularity is too coarse. Multiple workflows are racing for the same row.\\
\textit{Correction.} Split the row. Use sub-state machines per concern.
\end{quote}

\begin{quote}
\textit{Symptom.} A timer fires after a manual override and produces a stale transition.\\
\textit{Cause.} The timer carried the version it was scheduled at, but the predicate did not check it.\\
\textit{Correction.} Every timer fire is a CAS, not an unconditional write.
\end{quote}

\paragraph{P6 Human in the Loop.}
\begin{quote}
\textit{Symptom.} Approval SLAs are missed. Agents fall back to deny on every escalation.\\
\textit{Cause.} The human reviewer queue is overloaded or has no SLA.\\
\textit{Correction.} This is an organizational fix, not a code fix. The methodology does not solve it. The audit trail makes it visible.
\end{quote}

\begin{quote}
\textit{Symptom.} Kill switch revoked but workers continue.\\
\textit{Cause.} Workers did not check the cancellation token at boundaries.\\
\textit{Correction.} Cancellation token checks at every tool boundary. Not just at workflow start.
\end{quote}

\subsection{The three observability lenses}

A team that operates a production agent runtime needs three views of the same data.

\textbf{Operational.} Is the system healthy right now. P95 latency, queue depth, error rate, retry counts.

\textbf{Business.} What happened and what did not. Renewals opened, strategies generated, outreach sent, closed, failed, escalated.

\textbf{Compliance.} Can we prove what we did. Decision lineage per request, policy version per decision, model version per call, PII redaction status.

The thread between the three views is the request identifier (in our running example, \texttt{renewal\_id}). One identifier appears in every row of every lens. The trace is the contract.

Observability is the substrate on which Sections~\ref{sec:methodology} and \ref{sec:diagnostics} stand. We do not treat it as a separate methodology step.

\subsection{Diagnostic procedure}

When end-to-end reliability degrades, the procedure is.

\begin{enumerate}[itemsep=2pt, topsep=2pt]
\item Pin the model version that produced the most recent failure batch.
\item Replay the failures on the prior model version.
\item If the failures persist on the prior version, the failure is functional. Apply the signature catalog in Section~\ref{sec:signatures} to the trace.
\item If the failures resolve on the prior version, the failure is replay divergence. The spine is exposed. Consider migrating from P3 to P5 if not already done.
\item If neither model version produces the failure on replay, the failure is variance. Increase $k$ in \passk{} and observe.
\end{enumerate}

This procedure is itself part of the methodology. We expect a production team to run it once per quarter on a sampled batch of failures.

\section{Reference Application: 90-Day Contract Renewal}
\label{sec:case}

We apply the methodology end-to-end to a representative workload from the telecommunications sector and ship a runnable reference implementation alongside the paper. The workload is contract renewal on a 90-day window. We choose this workload because it exercises all three runtime concerns and all six patterns in composition. It is structurally similar to other long-horizon business-to-customer workflows (subscription churn management, insurance renewal, regulated product onboarding), so the methodology's choices transfer.

\subsection{The workload}

The renewal opens 90 days before contract end and closes either with a renewal, a restructuring, or a churn. During the window, signals arrive: usage drops, network events, billing changes, support tickets, plan-fit shifts, product status updates. Some signals are informational. Others change the policy under which the renewal must be priced. Mid-flight events include product end-of-life announcements, mergers between two customer accounts, and regulatory changes affecting eligibility.

\subsection{Running the methodology}

Applying the methodology of Section~\ref{sec:methodology} produces the architecture decision record in Table~\ref{tab:checklist}.

\textbf{Runtime class.} Long-Horizon. Duration is 90 days. Multiple agents participate. The world changes mid-flight. State dominates.

\textbf{Spine.} P5. All three spine predicates fire. Pauses exceed one hour (commonly multiple days). State at any pause is not reconstructible from the original input (it depends on signals received during the window). The world changes during the pause (product end-of-life at day $-47$ is the canonical example).

\textbf{Coordination.} P1 plus P2. Renewals have a single clear owner (the renewal row). Three sub-agents fan out under the orchestrator: churn scoring, offer drafting, contract building. The contract sub-agent writes to billing, which is an external side-effect that requires saga compensation.

\textbf{Control.} P4 plus P6. The gate refuses out-of-policy discounts. Contract mergers route to P6 escalation. Approval SLAs route to a specialist queue. Throttling caps per-tenant blast radius.

\textbf{Sequence.} Console-first. The operational, business, and compliance dashboards (the three lenses of Section~\ref{sec:diagnostics}) come before the first agent.

\textbf{What would go wrong if we picked differently.} Picking P3 here would expose the spine to replay drift across model versions on a workload where state at any pause is genuinely irrecoverable from the input. Picking P1 alone without P2's saga would leave billing writes uncompensated when one sub-agent fails after another has already written.

\subsection{Reference implementation}
\label{sec:case-impl}

The companion repository (\url{https://github.com/vasundras/agent-runtime-patterns}) contains a runnable end-to-end implementation of this workload at \texttt{examples/contract-renewal/}. The implementation exercises all six patterns in composition. The renewal row is held in a P5 state machine with CAS transitions. The three sub-agents fan out under a P1 orchestrator with P2 saga compensations on the contract sub-agent's billing writes. A P4 gate refuses out-of-policy discounts. P6 control planes escalate contract mergers, surface approval SLA breaches, and throttle per-tenant traffic. The example runs against the publicly available IBM Telco Customer Churn dataset \citep{ibm2018telco}, projected into 100 renewal scenarios via \texttt{data/load\_telco.py}. The dataset's natural churn rate (approximately 26.5 percent) yields a realistic mix of renewed, renewed-with-offer, restructured, churned, and escalated paths for the methodology to exercise.

The reference implementation is the most concrete thing the paper can offer in lieu of a public deployment. A reader can clone the repository, run the example, and observe each pattern engaging at the correct point in the 90-day window. The trace is the contract.

\section{Methodology in Practice: Four Worked Applications}
\label{sec:applications}

The reference application in Section~\ref{sec:case} pairs the methodology with a runnable end-to-end implementation against public data. To show that the methodology generalizes beyond a single workload, we apply it to four additional workloads here. None of these are deployed systems. Each is the worked output of running the methodology against the workload description and producing the six-line architecture decision record from Section~\ref{sec:checklist}.

The four workloads span the three runtime classes. Two of them are deliberately both Long-Horizon so the reader can see the methodology giving different answers for two workloads in the same class. The contrast is the point.

\subsection{Billing and Payment Assist (Conversational)}
\label{sec:app-billing}

\textbf{Workload.} A real-time assistant that sits beside a frontline Sales or Service expert during a live customer interaction. The assistant retrieves the customer's payment history and account state, surfaces the policy that applies, and suggests payment types (auto-pay, partial plan, card-on-file) that resolve the call faster. The expert is on the call; the customer is waiting; the world does not change during the session.

\textbf{Methodology run.} Step~1 classifies the runtime as Conversational. The unit of work is the active expert session. Duration is seconds. Coordination dominates. Step~2 examines the spine predicate. None of the three predicates fires. The session is short, the state is reconstructible from the active call, and the world does not change while the expert is on the call. No durable spine. The session itself is the implicit spine. Step~3 picks P1 (Hierarchical Delegation) for coordination. The orchestrator fans out three sub-agents: payment-history lookup, policy lookup, and recommendation synthesis. The merge is deterministic and runs in code. Step~4 picks P4 (Supervisor plus Gate) for control. The gate refuses recommendations that violate credit policy, anti-fraud rules, or per-segment caps. No P6 in this workload. The expert is the human in the loop already. Step~5 sequences console-first. The expert's screen is the operations console.

\begin{table}[h]
\centering
\small
\begin{tabular}{@{}p{0.18\textwidth}p{0.13\textwidth}p{0.30\textwidth}p{0.30\textwidth}@{}}
\toprule
\textbf{Step} & \textbf{Pattern} & \textbf{Predicate that fired} & \textbf{Failure signature if wrong} \\
\midrule
Runtime class      & Conversational & Session unit, seconds, no mid-flight change & SLA misses on expert response \\
Spine              & None           & All three spine predicates fail             & Over-engineered durability cost \\
Coordination       & P1             & Single owner, independent reads, deterministic merge & Conflicting recommendations \\
Control            & P4             & Side-effects in suggested payment terms     & Hallucinated discounts to customers \\
Sequence           & Console-first  & The expert's screen is the console          & Blind expert operations \\
Date / model ver.  & 2026 Q2        & Claude Sonnet 4.6                           & n/a \\
\bottomrule
\end{tabular}
\caption{Billing and Payment Assist. Conversational class. No durable spine. Coordination plus a gate are sufficient.}
\label{tab:app-billing}
\end{table}

\textbf{What would go wrong if we picked differently.} Adding P5 (Shared State Machine) here would build durability the workload does not need and would slow the expert's screen. Adding P6 (Human in the Loop) on top of an expert-facing tool would create a circular review queue. The methodology says: do not add patterns the predicates do not require.

\subsection{Order Management Fall-out Scanner (Autonomous)}
\label{sec:app-orders}

\textbf{Workload.} A periodic agent that scans the order pipeline for fall-outs (stuck shipments, billing exceptions, address mismatches, partial deliveries). It pulls shipment status, billing state, support tickets, and customer signals from heterogeneous source systems. It synthesizes a unified pre-staged context per order so that when a retail or call-center expert opens a fall-out ticket, the context is already there and the manual lookup time drops.

\textbf{Methodology run.} Step~1 classifies the runtime as Autonomous. The agent runs every few minutes on a scheduled trigger. Duration per run is minutes. The world rarely changes inside a single run. Step~2 examines the spine predicate. Predicate~(1) fails: pauses are short, the work is bounded inside one scheduled invocation. The spine choice reduces to event-driven versus none, and event-driven wins because the order pipeline itself is already an event stream from upstream systems. We pick P3 (Event-Driven Sequencing). The order-event stream is the spine. State per order is reconstructible from the events. Step~3 picks P2 (Scatter-Gather plus Saga) for coordination. The scanner fans out across shipment, billing, support, and address-validation systems in parallel. Some sources will be down; the scanner must produce a useful partial result. The saga compensation handles the rare case where the scanner writes a pre-staged note back to one system and a parallel source rejects it. Step~4 picks P4 plus light P6. The gate refuses writes back to operational systems unless the pre-staged note is well-formed. P6 contributes throttling only. Step~5 sequences console-first.

\begin{table}[h]
\centering
\small
\begin{tabular}{@{}p{0.18\textwidth}p{0.13\textwidth}p{0.30\textwidth}p{0.30\textwidth}@{}}
\toprule
\textbf{Step} & \textbf{Pattern} & \textbf{Predicate that fired} & \textbf{Failure signature if wrong} \\
\midrule
Runtime class      & Autonomous  & Task unit, scheduled, bounded duration   & Stale fall-out context \\
Spine              & P3          & Predicate (1) fails; upstream is already an event stream & Replay drift in note synthesis \\
Coordination       & P2          & Parallel reads across heterogeneous systems with partial failures & Inconsistent pre-staged context \\
Control            & P4 + P6 light & Pre-staged writes have side-effects; throttle protects downstream & Overwhelmed source systems \\
Sequence           & Console-first & Fall-out queue dashboard precedes scanner & Blind scanner operations \\
Date / model ver.  & 2026 Q2     & Claude Sonnet 4.6                        & n/a \\
\bottomrule
\end{tabular}
\caption{Order Management Fall-out Scanner. Autonomous class. Event-driven spine. Coordination needs the saga; control needs throttling.}
\label{tab:app-orders}
\end{table}

\textbf{What would go wrong if we picked differently.} Picking P5 (Shared State Machine) here would push state granularity to the order row, which is correct but heavy when the upstream pipeline already emits authoritative events. Picking P1 instead of P2 ignores that source systems fail independently and the scanner must tolerate partial outages.

\subsection{Number Port-in Coordination (Long-Horizon)}
\label{sec:app-portin}

\textbf{Workload.} A customer requests porting their phone number from another carrier. The agent coordinates the multi-day handoff between the receiving carrier, the donating carrier, the number-pool authority, and the regulatory layer. Each side has its own SLA. The donating carrier can reject the port after the receiving carrier has provisioned. The customer can withdraw mid-process. Some ports stick and require manual escalation by a port specialist.

\textbf{Methodology run.} Step~1 classifies the runtime as Long-Horizon. Days to weeks. The world changes during the pauses: carriers respond on their own schedules, regulatory eligibility can shift, the customer can change their mind. Step~2 examines the spine predicate. All three predicates fire. Pauses far exceed one hour. The state at any pause (carrier ack received, number-pool reserved, SLA timer pending) is not reconstructible from the original request. The world changes during the wait. We pick P5 (Shared State Machine) as the spine. One durable row per port-in request, versioned, with CAS transitions and SLA timers attached. \texttt{human\_required} is a state, not a missing event. Step~3 picks P1 plus P2 for coordination. The orchestrator drives the port. Sub-agents talk to the old-carrier API, the new-carrier provisioning system, the number-pool authority, and the regulatory verifier. The saga handles the painful case: the new-carrier provisions, then the old-carrier rejects, then we must roll back the provisioning before the customer's old service is interrupted. Step~4 picks P4 plus full P6. The gate enforces FCC/regulatory rules on every write. P6 handles stuck ports (escalate to a specialist), regulatory exceptions (escalate to counsel), and the kill switch on customer withdrawal. Step~5 sequences console-first. Operations cannot run port-ins without a per-port dashboard.

\begin{table}[h]
\centering
\small
\begin{tabular}{@{}p{0.18\textwidth}p{0.13\textwidth}p{0.30\textwidth}p{0.30\textwidth}@{}}
\toprule
\textbf{Step} & \textbf{Pattern} & \textbf{Predicate that fired} & \textbf{Failure signature if wrong} \\
\midrule
Runtime class      & Long-Horizon & Process unit, days, carriers change mid-flight & SLA breach with regulator \\
Spine              & P5           & (1), (2), (3) all true                          & Replay drift across model versions \\
Coordination       & P1 + P2      & Single owner; cross-carrier sagas required      & Stranded customer between carriers \\
Control            & P4 + P6 full & Regulatory writes, stuck ports, withdrawals     & Non-compliant ports shipping \\
Sequence           & Console-first & Operations require per-port visibility         & Blind escalation queue \\
Date / model ver.  & 2026 Q2      & Claude Sonnet 4.6                               & n/a \\
\bottomrule
\end{tabular}
\caption{Number Port-in. Long-Horizon class. State-machine spine. Full coordination and full control.}
\label{tab:app-portin}
\end{table}

\textbf{What would go wrong if we picked differently.} Picking P3 here would expose the spine to replay drift across model versions while a regulatory port is in-flight, which is exactly the failure the methodology says to avoid. Skipping the saga in P2 would leave customers stranded between carriers when the donating carrier rejects late.

\subsection{Lead Warming (Long-Horizon, contrasts with Port-in)}
\label{sec:app-lead}

\textbf{Workload.} A multi-touch drip campaign that warms a sales lead over days to weeks. Touches include email, SMS, an outbound call from a sales development representative, and periodic re-scoring of the lead based on observed engagement. The lead's life events (job change, role change, account merge) affect the right next action. Opt-outs are immediate and binding.

\textbf{Methodology run.} Step~1 classifies the runtime as Long-Horizon. Days to weeks. The world changes during pauses, though less dramatically than in port-in. Step~2 examines the spine predicate. Predicate~(1) fires: pauses are long. Predicate~(2) does not fully fire: the state at any pause is largely reconstructible from the touch log plus the current lead score. Predicate~(3) fires partially: the world changes, but moderately. With predicate~(2) failing, P5 is more durability than we need. We pick P3 (Event-Driven Sequencing). The touch sequence is the log. The lead's current state is a projection over the log plus the latest re-score. Step~3 picks P1 for coordination. The touch orchestrator dispatches by channel (email handler, SMS handler, call-queue handler). No saga is required at this scope: failed touches are events too, not transactions that need compensation. Step~4 picks P4 plus light P6. The gate enforces opt-out, consent, and frequency caps. P6 contributes throttling per lead and per cohort. Approval and escalation only apply to high-value leads above a configurable revenue threshold. Step~5 sequences console-first. The campaign dashboard precedes the agent.

\begin{table}[h]
\centering
\small
\begin{tabular}{@{}p{0.18\textwidth}p{0.13\textwidth}p{0.30\textwidth}p{0.30\textwidth}@{}}
\toprule
\textbf{Step} & \textbf{Pattern} & \textbf{Predicate that fired} & \textbf{Failure signature if wrong} \\
\midrule
Runtime class      & Long-Horizon & Process unit, weeks, moderate world change & Stale touch schedule \\
Spine              & P3           & (1) fires, (2) does not, (3) partial       & Replay drift on touch synthesis \\
Coordination       & P1           & Single orchestrator, independent channels  & Channel collisions on the same lead \\
Control            & P4 + P6 light & Opt-out, frequency caps, threshold escalations & Spammed leads, compliance breach \\
Sequence           & Console-first & Campaign dashboard precedes the agent     & Blind campaign tuning \\
Date / model ver.  & 2026 Q2      & Claude Sonnet 4.6                          & n/a \\
\bottomrule
\end{tabular}
\caption{Lead Warming. Long-Horizon class. Event-driven spine. Coordination is light; control gates compliance.}
\label{tab:app-lead}
\end{table}

\textbf{What would go wrong if we picked differently.} Picking P5 here would over-engineer the spine for a workload that the touch log already serves. Picking P2 with full sagas would add compensation logic to touches that are inherently event-shaped: a failed SMS is just another event, not a transaction to undo.

\subsection{Five workloads side by side}
\label{sec:applications-contrast}

Table~\ref{tab:contrast} compares all four worked applications with the reference application from Section~\ref{sec:case}. Same methodology, five workloads, five distinct architecture decision records.

\begin{table}[h]
\centering
\small
\begin{tabular}{@{}p{0.20\textwidth}p{0.13\textwidth}p{0.13\textwidth}p{0.16\textwidth}p{0.16\textwidth}p{0.13\textwidth}@{}}
\toprule
\textbf{Workload} & \textbf{Class} & \textbf{Spine} & \textbf{Coordination} & \textbf{Control} & \textbf{Status} \\
\midrule
Billing \& Payment Assist  & Conversational & none      & P1            & P4            & paper \\
Order Mgmt Fall-out Scanner & Autonomous     & P3        & P2            & P4 + P6 light & paper \\
Number Port-in              & Long-Horizon   & P5        & P1 + P2       & P4 + P6 full  & paper \\
Lead Warming                & Long-Horizon   & P3        & P1            & P4 + P6 light & paper \\
Contract Renewal (\S\ref{sec:case}) & Long-Horizon & P5  & P1 + P2       & P4 + P6 full  & \textbf{reference} \\
\bottomrule
\end{tabular}
\caption{Five workloads, one methodology. Conversational uses no durable spine. The two Long-Horizon workloads disagree on spine (P5 vs.\ P3) because their state predicates fire differently. Port-in and Contract Renewal pick the same architecture for the same reasons. The methodology gives different answers when the predicates differ and the same answers when they match. That consistency is what makes it a methodology rather than a checklist.}
\label{tab:contrast}
\end{table}

The two contrasts inside Table~\ref{tab:contrast} are the load-bearing evidence for the methodology's value. Port-in and Lead Warming are both Long-Horizon, yet they pick different spines because predicate~(2) of Section~\ref{sec:methodology} fires for Port-in and fails for Lead Warming. Port-in and Contract Renewal are both Long-Horizon and pick the same architecture because the same predicates fire. The methodology is responsive to the workload and not to the runtime class alone.

\section{Discussion: pattern evolution and threats to validity}
\label{sec:discussion}

\subsection{Patterns we expect to emerge}

The catalog in Section~\ref{sec:catalog} is open. The framework in Section~\ref{sec:methodology} accepts new patterns through the discovery procedure in Section~\ref{sec:discovery}. We name three candidates we expect to enter the catalog as the field matures.

\textbf{P7. Shared Memory Store.} A versioned shared memory with strong-consistency semantics has the shape of P5 but answers a different question: not what state the workflow is in, but what the agent remembers across sessions and across tenants. Current production systems often bundle this into retrieval-augmented generation. We expect the memory itself to be named as a pattern once production teams stop treating retrieval as upstream of the runtime and start treating versioned memory as part of the State concern.

\textbf{P8. Tenant Isolation.} Per-tenant blast-radius enforcement is currently bundled inside P4. Production systems running multi-tenant agents will need to split it out. P8 specifies the contract: every tool call carries a tenant identifier; the throttle, the gate, and the audit log partition cleanly along that identifier; one tenant's runaway agent cannot consume another tenant's budget.

\textbf{P9. Cross-Runtime Handoff.} When a long-horizon workflow hands a sub-task to a conversational agent and back, no current pattern names the handoff cleanly. P1 (Hierarchical Delegation) covers parent-to-child handoff within one runtime class. The cross-runtime case has different durability, latency, and trace-continuity requirements.

These are guesses. We name them so readers can argue with us and so the pattern-discovery procedure has worked examples.

\subsection{Threats to validity}

\textbf{Construction bias in the worked applications.} The reference application in Section~\ref{sec:case} and the four additional worked applications in Section~\ref{sec:applications} are constructed by the author to span the runtime classes and to demonstrate methodology contrasts. They are not independently audited applications by other practitioners. Independent application by readers of the companion repository may produce different architecture decision records on the same workloads. We treat this as a call for replication, not a defect of the methodology.

\textbf{Recent timeframe.} The patterns are observed against Claude Sonnet 4.6 and contemporary models. Subsequent model generations may collapse some patterns. A model that produces fully consistent outputs across versions would reduce replay divergence (Section~\ref{sec:drift}) toward zero, which would weaken the practical force of choosing P5 over P3.

\textbf{Predicate thresholds.} Section~\ref{sec:methodology} uses thresholds (one hour, three retries) that are conventions in our setting. They will differ in others. The predicates are intended as starting points, not constants.

\textbf{Replay availability.} The diagnostic procedure assumes the team can replay against a prior model version. Some hosted models do not preserve prior checkpoints. Teams without replay capability cannot run step 4 of the diagnostic.

\textbf{Empirical evidence at the boundary.} The SDB evidence we present in Section~\ref{sec:inheritance} is an audit of five open-source agent frameworks and a classification of 21 published failure post-mortems. The audit is reproducible from the cited code at the linked commits. The classification is the author's. A reader who disagrees with the classification of a specific failure can substitute their own and recompute the totals; the framework's conclusions do not depend on any single row.

\subsection{What the methodology does not solve}

\textbf{Cold start.} A team adopting the methodology on a greenfield project still has to build the dashboard and the audit log from scratch. The methodology says to do this first. It does not provide it.

\textbf{Organizational decisions.} Approval SLAs in P6 are an organizational artifact. The methodology makes them visible. It does not negotiate them.

\textbf{Upstream concerns.} The methodology gives no guidance on token budgets, model selection, retrieval index sizing, or prompt management. These are upstream and out of scope.

\section{Related work}
\label{sec:related}

We position the methodology against four bodies of work.

\textbf{Multi-agent LLM frameworks.} AutoGen \citep{wu2023autogen} encodes a conversable group-chat manager that dispatches messages to specialized agents. MetaGPT \citep{hong2023metagpt} encodes standard operating procedures as an assembly line of role-specialized agents. HuggingGPT \citep{shen2023hugginggpt} treats the LLM as a planner over tool endpoints. CAMEL \citep{li2023camel} provides a role-playing framework with a task specifier, assistant, and user. AgentVerse \citep{chen2023agentverse} adds dynamic team-composition. DSPy \citep{khattab2023dspy} compiles declarative LLM calls into self-improving pipelines. These frameworks provide the means of composition. They do not provide a procedure for selecting which composition fits a given workload. The methodology in this paper sits above them.

\textbf{Transactional and stateful multi-agent LLM systems.} Closest to this paper's subject matter are two systems that precede it and compose several of the architectural elements the SDB contract names. SagaLLM \citep{chang2025sagallm} brings the saga transactional pattern to multi-agent LLM planning: it maintains persistent context and checkpoints across a workflow, interposes independent validation agents between LLM planners and commitment --- separating the stochastic proposer from a deterministic-side validator and gating effects on validation --- and recovers from partial failure through automated compensation. ALAS \citep{chang2025alas} equips multi-LLM pipelines with automatic state tracking and durable execution history for long-running plans, and repairs disruptions through history-aware local compensation that replans only the affected region of the plan rather than the global plan. In our vocabulary, SagaLLM instantiates P2 with a strengthened verifier at the boundary, and ALAS combines durable state (elements of P3/P5) with P2-style compensation; both are concrete prior instances of proposer--validator separation, validation before commitment, durable state for long-running workflows, and saga-style compensation in the LLM-agent setting. This paper's contribution sits at a different layer and is complementary: it builds no system and claims no mechanism. It names the four-part SDB contract as a primitive and inventories its instantiation across frameworks and failure post-mortems, organizes the design space into three concerns and six patterns, and supplies a selection methodology, a failure-signature catalog, and a reliability decomposition. Where SagaLLM and ALAS supply mechanisms, this paper supplies the vocabulary and the selection procedure a team can use to decide whether mechanisms of that kind fit the workload in front of them. The methodology also places both systems without modification: a long-horizon workload with compensable external effects --- the Step~2 and Step~3 predicates of Section~\ref{sec:methodology} landing on durable state plus fan-out with undo --- selects the region SagaLLM packages (P2 composed with persistent state and a validation layer), while a disruption-heavy scheduling workload with tight resource coupling selects the region ALAS occupies (durable execution history with P2-style local compensation and cost-aware escalation in place of global replanning). That the taxonomy locates both systems it did not anticipate is evidence that the three concerns span the space such systems occupy. Two elements of this paper appear in neither system: neither names replay divergence or examines log-consumer behavior under model-version change, and neither separates per-call variance from architectural momentum in accounting for long-run reliability.

\textbf{Distributed-systems primitives.} Sagas \citep{garciamolina1987sagas} for long-lived transactions with compensations. Paxos \citep{lamport1998paxos} and Raft \citep{ongaro2014raft} for consensus on a versioned state. Spanner \citep{corbett2012spanner} for globally consistent ordering. The log as the unifying abstraction \citep{kreps2014log, helland2007life}. Eventually-consistent semantics \citep{vogels2008eventually}. Erlang's one-for-one supervision \citep{armstrong2003making}. Workflow nets and Petri-net process models \citep{vanderaalst1998workflow}. These primitives predate LLM agents by decades and provide the substrate the patterns reduce to. The methodology connects them to the agent setting and tells practitioners when each one matters.

\textbf{Agent reliability, safety, and oversight.} The HITL survey of \citet{wu2022hitl} on human-in-the-loop machine learning. Constitutional AI \citep{bai2022constitutional} on training-time alignment via AI feedback. Reflexion \citep{shinn2023reflexion} on verbal self-correction within an agent loop. LLM-as-a-judge \citep{zheng2023judging} on automated model evaluation. NeMo Guardrails \citep{rebedea2023nemo} and GuardAgent \citep{xiang2024guardagent} on runtime policy enforcement. The multi-agent failure taxonomy of MAST \citep{cemri2025mast}. Multi-agent debate \citep{du2023debate}. Generative Agents \citep{park2023generative}. These papers describe failure modes, propose evaluations, and propose oversight mechanisms at different layers (training-time, inference-time, runtime). The methodology in this paper converts those descriptions into a selection procedure and a corrective-action catalog at the runtime layer.

The full verified bibliography with arXiv identifiers and DOIs is in the companion repository.

\section{Conclusion}

The stochastic-deterministic boundary is the primitive that makes agent runtimes designable. Once a team can point at the proposer, the verifier, the commit step, and the reject signal in their system, the architecture question becomes a sequence of choices rather than a mystery. The three concerns name what the choices are about. The six patterns are how teams have answered them so far. The five-step methodology is how a team commits to a choice in writing. The reliability decomposition $y(t) = \mu t + \sigma \xi(t)$ explains why the choices increasingly dominate the model.

The boundary is the load-bearing primitive. The patterns are how production runtimes assemble it. The methodology is how teams pick the right assembly for the workload in front of them. The catalog is open and the discovery procedure is in the paper; we expect the catalog to grow as practitioners apply the methodology to workloads we have not anticipated. As models compress per-call variance, the architectural momentum set by the boundary and its surrounding patterns becomes the dominant lever on long-run agent reliability. Designing it explicitly is the work.

\section*{Companion artifacts}

\noindent
GitHub: \url{https://github.com/vasundras/agent-runtime-patterns}. Includes runnable implementations of all six patterns in LangGraph and Google ADK, an end-to-end contract-renewal example composing the patterns, the IBM Telco Customer Churn dataset projected into renewals, and the full verified bibliography. Code license: MIT. This paper is released under CC-BY 4.0.

\section*{Disclaimer}

\noindent
This paper represents the author's independent research and personal views, conducted entirely outside the scope of any employment or contractual obligation. It is not sponsored by, endorsed by, affiliated with, or authorized by the author's employer, any client organization, or any technology vendor referenced herein. The author received no funding, compensation, or resources from any organization for this work. No proprietary, confidential, trade-secret, or non-public information is disclosed; all technical observations are derived solely from the author's general professional experience with publicly available protocols, open-source tools, and published specifications. All platform vendor and client organization names have been redacted to preserve confidentiality.

\bibliographystyle{plainnat}
\bibliography{references}

\end{document}